\documentclass[journal,twoside,web]{ieeecolor2}
\usepackage{generic}
\usepackage{cite}
\usepackage{amsmath,amssymb,amsfonts}
\usepackage{algorithmic,algorithm}
\usepackage{graphicx,multirow}
\usepackage{textcomp}
\usepackage{booktabs}
\usepackage[hidelinks]{hyperref}
\def\BibTeX{{\rm B\kern-.05em{\sc i\kern-.025em b}\kern-.08em
		T\kern-.1667em\lower.7ex\hbox{E}\kern-.125emX}}
\begin{document}
\title{KDHNet: A Knowledge-Driven Hybrid Network for Limited-Calibration SSVEP Decoding}
\author{Zhentao~He, Ziwei~Wang$^{\ast}$, \IEEEmembership{Student Member,~IEEE}, and Dongrui~Wu$^{\ast}$, \IEEEmembership{Fellow,~IEEE}
\thanks{This research was supported by the National Natural Science Foundation of China under Grants (625B2077 and 62525305), and in part by the Open Fund of Intelligent Control Laboratory 2024-ZKSYS-KF02-04.}
\thanks{Z. He, Z. Wang, and D. Wu are with School of Artificial Intelligence and Automation, Huazhong University of Science and Technology, Wuhan 430074, China.}
\thanks{*Corresponding Authors: Ziwei Wang (vivi@hust.edu.cn) and Dongrui Wu (drwu09@gmail.com).}
}

\maketitle

\begin{abstract}
\textit{Objective:} Steady-state visual evoked potential (SSVEP)-based brain-computer interfaces (BCIs) achieve high information transfer rates but often require sufficient subject-specific calibration data for reliable decoding. Analytical spatial filters are effective with limited data, but their linear formulations constrain representation capacity. Deep models learn more flexible representations but are prone to overfitting when calibration data are scarce. \textit{Methods:} We propose KDHNet, which combines knowledge-driven analytical processing with data-driven representation learning. After filter-bank preprocessing, cascaded task-discriminant component analysis (TDCA) and task-related component analysis (TRCA) extract class-discriminative and trial-reproducible components. Learnable sub-band weighting then fuses spectral information, and an SSVEPformer decoder learns temporal and spectral representations for classification. For inter-subject decoding, a target-guided least-squares transformation aligns source trials using one labeled target trial per class. \textit{Results:} Extensive experiments were performed on three public datasets, covering intra- and inter-subject protocols, five analysis windows, and multiple calibration ratios. In intra-subject evaluation, KDHNet achieved the highest mean accuracy and information transfer rate in 14 of 15 window settings and all ten calibration-ratio settings. In inter-subject evaluation, KDHNet obtained the best or second-best performance, and the gains were generally larger with less calibration data. Ablation studies indicated that the analytically estimated TDCA projection performed better when fixed than when optimized from analytical or random initialization. \textit{Conclusion:} Combining fixed knowledge priors with trainable neural modules is effective for limited-calibration SSVEP decoding. \textit{Significance:} This framework helps reduce the calibration burden of SSVEP decoding and supports more practical, robust, and plug-and-play BCIs.
\href{https://github.com/2665023991/KDHNet}{KDHNet code} is publicly available.
\end{abstract}


\section{Introduction}

\IEEEPARstart{B}{rain}-computer interfaces (BCIs) establish a direct communication pathway between the brain and external devices \cite{Wolpaw2002,Rosenfeld2017} and electroencephalography (EEG) is the most widely used non-invasive modality for BCIs, with applications ranging from device control \cite{wang2025mvcnet} to cognitive monitoring \cite{zheng2017identifying} and clinical assessment \cite{wang2025cst}. Among various BCI paradigms, steady-state visual evoked potential (SSVEP) is one of the most effective tasks, offering high information transfer rates (ITRs), short user training \cite{SSVEPReview2021}, and simple stimulation protocols \cite{DecodingReview2022}. Repetitive visual stimulation evokes oscillatory EEG responses at the stimulus frequency and its harmonics, so the attended target can be identified from occipital EEG, which has supported high-speed spellers and assistive-control systems \cite{bin2011high}. In practice, however, reliable decoding still depends on subject-specific calibration data, and lengthy calibration sessions increase visual fatigue and hinder plug-and-play deployment \cite{Huang2023DGTransfer,Chiang2021LST}. Reducing this calibration burden, either by decoding with limited training trials or by transferring knowledge from auxiliary subjects, is therefore central to enhancing SSVEP decoding, yet it remains difficult because EEG has a low signal-to-noise ratio (SNR) and large inter-session and inter-subject variability \cite{Huang2023DGTransfer}.

Existing SSVEP decoders can be broadly grouped into knowledge-driven and data-driven approaches \cite{DecodingReview2022,Xu2023DLSurvey}. Knowledge-driven decoders exploit prior neurophysiological structure. Canonical correlation analysis (CCA) and its filter-bank extension (FBCCA) match EEG responses to sinusoidal references in carefully designed sub-bands \cite{CCA,FBCCA,bin2009online}. Later variants, such as multiway CCA \cite{zhang2011multiway}, phase-constrained CCA \cite{pan2011enhancing}, the multivariate synchronization index \cite{MSI2014}, and the sum of squared correlations \cite{SSCOR2019}, further improve frequency recognition. Task-related component analysis (TRCA) \cite{tanaka2013task,TRCA} and its similarity-constrained variant \cite{sun2021similarity} learn spatial filters that maximize trial reproducibility, while task-discriminant component analysis (TDCA) adds class discrimination \cite{Liu2021TDCA}. These algorithms are interpretable and remain reliable when only a few calibration trials are available. However, their analytically derived linear transformations have limited capacity to capture complex discriminative patterns.

Data-driven decoders instead learn discriminative representations directly from EEG in an end-to-end manner. Compact CNNs \cite{EEGNet,Waytowich2018}, filter-bank and correlation-based networks \cite{Ding2021FBtCNN,CCNN,ConvCA}, recurrent convolutional models \cite{SSVEPNet}, and more recent Transformer-based models \cite{wang2025dbconformer,SSVEPformer,Guney2022,song2022eeg} have improved SSVEP decoding by jointly optimizing feature extraction and classification \cite{Xu2023DLSurvey}. However, deep models rely on sufficient labeled data and tend to become unstable when only a few subject-specific calibration trials are available, and their learned representations are less interpretable than analytical spatial filters.

Two lines of work aim to bridge these approaches. Hybrid decoders inject domain knowledge into neural networks. For example, TRCA-Net applies TRCA-derived spatial filtering before a CNN \cite{Deng2023TRCANet}, and prediction-level fusion of TRCA-style decoding and deep networks has also proven beneficial \cite{Wei2025eTRCASbCNN}. In parallel, inter-subject or cross-dataset transfer lowers the target-calibration requirement by aligning source data to the target \cite{wu2020transfer,wang2026ckd}, for example through least-squares transformation (LST) \cite{Chiang2021LST}, domain generalization \cite{Huang2023DGTransfer}, source knowledge distillation \cite{wang2026ckd}, or source selection \cite{Wang2023TASA}. Despite this progress, several challenges remain in limited-calibration SSVEP decoding:
\begin{itemize}
	\item The sample efficiency of analytical priors and the representation power of deep learning decoders are seldom exploited jointly, as the two families are typically used in isolation.
	\item Existing hybrids rely on CNN backbones or shallow prediction-level fusion and do not couple task-discriminative spatial priors with Transformer-based sequence modeling.
	\item The intra-subject and inter-subject limited-calibration regimes are usually served by separate pipelines rather than a single decoder.
\end{itemize}

To address these gaps, we propose KDHNet, a knowledge-driven hybrid network for limited-calibration SSVEP decoding. KDHNet cascades a closed-form TD-RCA (TDCA followed by TRCA) knowledge front-end with an SSVEPformer decoder. The front-end enhances task-discriminative and reproducible SSVEP components through analytical spatial filtering, and the Transformer models higher-order temporal-spectral dependencies of the resulting knowledge-enhanced features. A single decoder serves both limited-calibration regimes, the intra-subject setting and the inter-subject setting, where LST alignment first adapts source data to the target subject.

The main contributions of this work are summarized as follows:
\begin{itemize}
	\item We propose KDHNet, a knowledge-driven hybrid network for limited-calibration SSVEP decoding that couples a closed-form TD-RCA spatial-filtering front-end with a Transformer decoder, combining the sample efficiency and interpretability of knowledge-driven spatial filtering with the representation power of deep sequence modeling.
	\item We develop a unified framework for intra- and inter-subject decoding with limited calibration data. The same architecture is trained using a few target-subject trials in the intra-subject setting, whereas target-guided LST incorporates aligned source trials into inter-subject training.
	\item We evaluate KDHNet on three public benchmarks under intra- and inter-subject scenarios across a range of time windows and calibration ratios. Results showed that KDHNet outperformed representative traditional, deep-learning, and hybrid baselines in 14 of 15 intra-subject window settings, all ten calibration-ratio settings, and two of the three inter-subject benchmarks, with the largest gains under the scarcest calibration. A component-wise ablation further explained this behavior, showing that the analytical front-end contributes as a structural prior and enhances the end-to-end training.
\end{itemize}

\section{Method}\label{sec:methods}

\subsection{Problem Formulation}

Let $\mathbf{X} \in \mathbb{R}^{N_c \times T}$ denote an EEG trial, where $N_c$ and $T$ denote the number of channels and time samples, respectively, and let $y \in \{1,2,\ldots,N_f\}$ denote the corresponding stimulus label. For the $u$-th subject, the subject-specific dataset is $\mathcal{D}_{u}=\{(\mathbf{X}_{u,i}, y_{u,i})\}_{i=1}^{N_u}$.

This study considers two limited-calibration SSVEP decoding scenarios. In the intra-subject scenario, the training and test trials are drawn from the same subject, and only a small portion of subject-specific trials is used for model calibration. In the inter-subject scenario, the leave-one-subject-out (LOSO) evaluation is adopted. One subject is treated as the target domain, the remaining subjects constitute the source domain, and a small target calibration set $\mathcal{D}_{t}^{\mathrm{cal}}$ is available from the target subject. The remaining target-subject trials are used as the test set $\mathcal{D}_{t}^{\mathrm{te}}$.

The same KDHNet decoder is used in both scenarios. The difference lies in the construction of the training set. For intra-subject decoding, KDHNet is trained directly on the limited subject-specific calibration trials. For inter-subject decoding, the source-domain data are first aligned to the target subject by LST using $\mathcal{D}_{t}^{\mathrm{cal}}$, and the aligned source data are then combined with the target calibration data for downstream training.

\begin{table}[t]
\caption{Summary of symbols used in this paper.}
\label{tab:notation_summary}
\centering
{\renewcommand{\arraystretch}{1.2}
\begin{tabular}{ll}
\toprule
Symbol & Definition \\
\midrule
$N_c$ & Number of EEG channels \\
$T$ & Number of time samples \\
$N_f$ & Number of stimulus classes \\
$N_m$ & Number of filter-bank sub-bands \\
$\mathbf{X}$ & Single-trial EEG segment \\
$y$ & Label of the EEG trial \\
$\mathcal{D}_{u}$ & Data of the $u$-th subject \\
$\mathcal{D}_{t}^{\mathrm{cal}}$ & Target-subject calibration set in LOSO \\
$\mathcal{D}_{t}^{\mathrm{te}}$ & Target-subject test set in LOSO \\
$\mathbf{S}_{v}^{f}$ & Source template of class $f$ for source subject $v$ \\
$\mathbf{T}_{t}^{f}$ & Target calibration template of class $f$ \\
$\mathbf{A}_{v}^{f}$ & Class-wise LST matrix from source subject $v$ to target $t$ \\
$\mathbf{X}^{(m)}$ & EEG trial in the $m$-th sub-band \\
$\mathbf{W}_{\mathrm{TD}}^{(m)}$ & TDCA projection matrix in the $m$-th sub-band \\
$\widehat{\mathbf{Z}}_{f,h}^{(m)}$ & Transformed TDCA feature before TRCA \\
$\psi$ & Batch normalization and PReLU parameters \\
$\mathbf{w}_{\mathrm{TR},f}^{(m)}$ & TRCA spatial filter of class $f$ in the $m$-th sub-band \\
$\mathbf{R}_{h}^{(m)}$ & TD-RCA feature of the $m$-th sub-band \\
$\alpha_m$ & Learnable weight of the $m$-th sub-band \\
$\theta$ & Parameters of the SSVEPformer decoder \\
$\mathbf{p}_{h}$ & Posterior probability vector predicted by KDHNet \\
\bottomrule
\end{tabular}
}
\end{table}

\subsection{Overview of KDHNet}

\begin{figure*}[t]
	\centering
	\includegraphics[width=\textwidth]{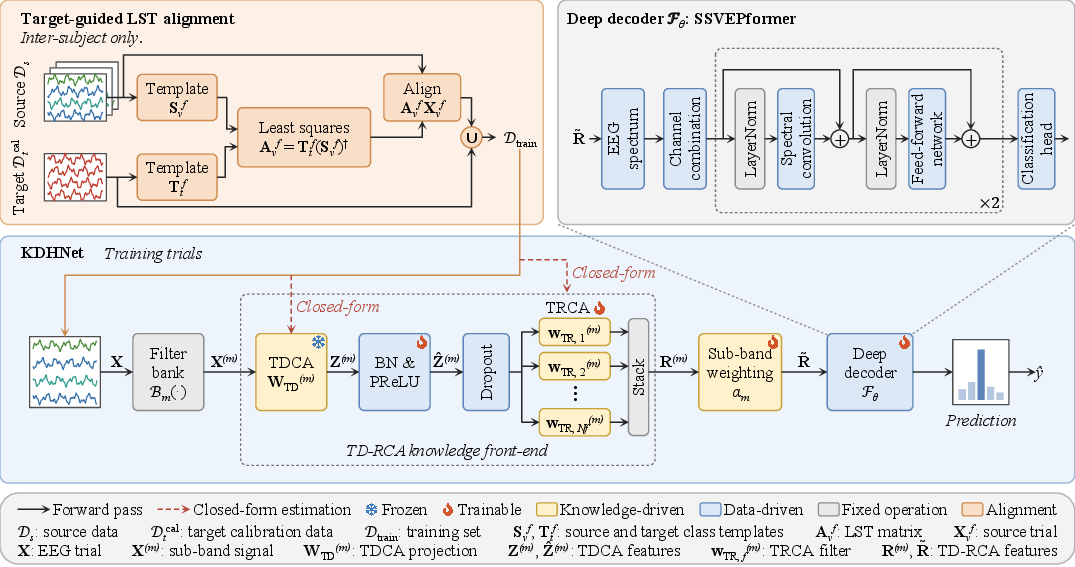}
	\caption{Overview of KDHNet. Bottom: the decoding pipeline, consisting of filter-bank preprocessing, the TD-RCA knowledge front-end, learnable sub-band weighting, and a deep decoder. The TDCA projections are estimated in closed form on the training set and kept frozen, whereas the TRCA filters and sub-band weights are initialized analytically and then trained end-to-end together with the decoder. Top left: target-guided LST alignment, which constructs the training set in the inter-subject setting. Top right: the SSVEPformer decoder used in this work. The symbols are defined in the legend, and the indices $v$, $t$, $f$, and $m$ denote the source subject, target subject, class, and sub-band, respectively.}
	\label{fig:kdhnet_architecture}
\end{figure*}

The overall architecture of KDHNet is illustrated in Fig.~\ref{fig:kdhnet_architecture}. Once the scenario-specific training set has been constructed, each trial is processed by filter-bank preprocessing, a TD-RCA knowledge front-end, learnable sub-band weighting, and an SSVEPformer decoder. In this way, KDHNet embeds prior knowledge into the decoding pipeline through the TD-RCA front-end and then performs deep temporal-spectral decoding with Transformer-based feature learning.

\subsection{Filter-Bank Preprocessing}
SSVEP responses are concentrated around the stimulation frequency and its harmonics. To exploit this property, each EEG trial is decomposed into multiple frequency sub-bands:
\begin{equation}
	\mathbf{X}^{(m)} = \mathcal{B}_m(\mathbf{X}), \qquad m = 1,2,\ldots,N_m,
	\label{eq:filterbank}
\end{equation}
where $\mathcal{B}_m(\cdot)$ denotes the $m$-th band-pass filter and $\mathbf{X}^{(m)} \in \mathbb{R}^{N_c \times T}$ is the corresponding sub-band signal.

This preprocessing step allows the model to capture complementary harmonic information and suppress broadband noise before knowledge-driven feature extraction.

\subsection{TD-RCA Knowledge Front-End}

The knowledge-driven front-end of KDHNet cascades TDCA and TRCA. The TDCA stage first extracts class-discriminative spatial projections, and the TRCA stage enhances the reproducibility of task-related components. The output of TDCA is further processed by batch normalization and a PReLU activation before TRCA.

\subsubsection{TDCA stage}
For the $m$-th sub-band, let $\mathbf{X}_{f,h}^{(m)} \in \mathbb{R}^{N_c \times T}$ denote the $h$-th training trial of class $f$, where $f=1,2,\ldots,N_f$. The class mean and global mean are defined as
\begin{equation}
	\bar{\mathbf{X}}_{f}^{(m)} = \frac{1}{n_f}\sum_{h=1}^{n_f}\mathbf{X}_{f,h}^{(m)},
\end{equation}
\begin{equation}
	\bar{\mathbf{X}}^{(m)} = \frac{1}{\sum_{f=1}^{N_f} n_f}\sum_{f=1}^{N_f}\sum_{h=1}^{n_f}\mathbf{X}_{f,h}^{(m)}.
\end{equation}

The between-class and within-class scatter matrices are computed as
\begin{equation}
	\mathbf{S}_{b}^{(m)} = \sum_{f=1}^{N_f} n_f
	\left(\bar{\mathbf{X}}_{f}^{(m)} - \bar{\mathbf{X}}^{(m)}\right)
	\left(\bar{\mathbf{X}}_{f}^{(m)} - \bar{\mathbf{X}}^{(m)}\right)^{\mathrm T},
\end{equation}
\begin{equation}
	\mathbf{S}_{w}^{(m)} = \sum_{f=1}^{N_f}\sum_{h=1}^{n_f}
	\left(\mathbf{X}_{f,h}^{(m)} - \bar{\mathbf{X}}_{f}^{(m)}\right)
	\left(\mathbf{X}_{f,h}^{(m)} - \bar{\mathbf{X}}_{f}^{(m)}\right)^{\mathrm T}.
\end{equation}

The TDCA projection matrix is obtained by
\begin{equation}
	\mathbf{W}_{\mathrm{TD}}^{(m)}
	=
	\arg\max_{\mathbf{W}}
	\frac{\mathrm{tr}\left(\mathbf{W}^{\mathrm T}\mathbf{S}_{b}^{(m)}\mathbf{W}\right)}
	{\mathrm{tr}\left(\mathbf{W}^{\mathrm T}(\mathbf{S}_{w}^{(m)}+\epsilon\mathbf{I})\mathbf{W}\right)},
	\label{eq:tdca}
\end{equation}
where $\epsilon$ is a small regularization constant. The TDCA-projected feature is
\begin{equation}
	\mathbf{Z}_{f,h}^{(m)} = \left(\mathbf{W}_{\mathrm{TD}}^{(m)}\right)^{\mathrm T}\mathbf{X}_{f,h}^{(m)}.
\end{equation}

Then, the feature is transformed as
\begin{equation}
	\widehat{\mathbf{Z}}_{f,h}^{(m)}
	=
	\phi\!\left(\mathrm{BN}\!\left(\mathbf{Z}_{f,h}^{(m)}\right)\right),
	\label{eq:bn_prelu}
\end{equation}
where $\mathrm{BN}(\cdot)$ denotes batch normalization and $\phi(\cdot)$ denotes the PReLU activation. During end-to-end training, dropout is further applied to $\widehat{\mathbf{Z}}_{f,h}^{(m)}$ before the trainable TRCA layer for regularization.

\subsubsection{TRCA stage}
For class $f$ in sub-band $m$, the TRCA spatial filter is estimated by
\begin{equation}
	\mathbf{w}_{\mathrm{TR},f}^{(m)}
	=
	\arg\max_{\mathbf{w}}
	\frac{\mathbf{w}^{\mathrm T}\mathbf{S}_{\mathrm{TR},f}^{(m)}\mathbf{w}}
	{\mathbf{w}^{\mathrm T}\mathbf{Q}_{\mathrm{TR},f}^{(m)}\mathbf{w}},
	\label{eq:trca}
\end{equation}
where
\begin{equation}
	\mathbf{S}_{\mathrm{TR},f}^{(m)}
	=
	\sum_{\substack{h,h'=1\\h\neq h'}}^{n_f}
	\widehat{\mathbf{Z}}_{f,h}^{(m)}
	\left(\widehat{\mathbf{Z}}_{f,h'}^{(m)}\right)^{\mathrm T},
\end{equation}
\begin{equation}
	\mathbf{Q}_{\mathrm{TR},f}^{(m)}
	=
	\sum_{h=1}^{n_f}
	\widehat{\mathbf{Z}}_{f,h}^{(m)}
	\left(\widehat{\mathbf{Z}}_{f,h}^{(m)}\right)^{\mathrm T}.
\end{equation}

Let $\mathbf{Z}_{\mathrm{in},f,h}^{(m)}$ denote the feature entering the TRCA layer:
\begin{equation}
\mathbf{Z}_{\mathrm{in},f,h}^{(m)}
=
\begin{cases}
\operatorname{Dropout}\!\left(\widehat{\mathbf{Z}}_{f,h}^{(m)}\right), & \text{training},\\
\widehat{\mathbf{Z}}_{f,h}^{(m)}, & \text{otherwise},
\end{cases}
\end{equation}
where the second case applies to TRCA initialization and evaluation. The TRCA response is then computed as
\begin{equation}
\mathbf{r}_{f,h}^{(m)}
=
\left(\mathbf{w}_{\mathrm{TR},f}^{(m)}\right)^{\mathrm T}
\mathbf{Z}_{\mathrm{in},f,h}^{(m)}.
\label{eq:trca_response}
\end{equation}

The responses from all classes are stacked as
\begin{equation}
	\mathbf{R}_{h}^{(m)}=
	\begin{bmatrix}
		\mathbf{r}_{1,h}^{(m)}\\
		\mathbf{r}_{2,h}^{(m)}\\
		\vdots\\
		\mathbf{r}_{N_f,h}^{(m)}
	\end{bmatrix}.
	\label{eq:stack}
\end{equation}

\subsection{Sub-Band Weighting}

After TD-RCA processing, the band-wise features are fused by a learnable weighting module. Let $\mathbf{R}_{h}^{(m)}$ denote the TD-RCA feature of the $m$-th sub-band. The weighted feature is defined as
\begin{equation}
	\widetilde{\mathbf{R}}_{h}^{(m)} = \alpha_m \mathbf{R}_{h}^{(m)},
	\label{eq:weighting}
\end{equation}
where $\alpha_m$ is the learnable coefficient of the $m$-th sub-band. In KDHNet, the sub-band coefficients are initialized according to
\begin{equation}
	\alpha_m^{(0)} = m^{-a} + b,
	\label{eq:alpha_init}
\end{equation}
where $a$ and $b$ are fixed initialization constants.

The fused representation is
\begin{equation}
	\widetilde{\mathbf{R}}_{h}=
	\begin{bmatrix}
		\widetilde{\mathbf{R}}_{h}^{(1)}\\
		\widetilde{\mathbf{R}}_{h}^{(2)}\\
		\vdots\\
		\widetilde{\mathbf{R}}_{h}^{(N_m)}
	\end{bmatrix}.
	\label{eq:fused}
\end{equation}

This design preserves complementary spectral information from different sub-bands while allowing the network to adapt their relative contribution to the decoding task.

\subsection{SSVEPformer Decoder}

The fused TD-RCA feature $\widetilde{\mathbf{R}}_{h}$ is fed into an SSVEPformer decoder for final classification:
\begin{equation}
	\mathbf{p}_{h} = \mathcal{F}_{\theta}(\widetilde{\mathbf{R}}_{h}),
	\label{eq:decoder}
\end{equation}
where $\mathcal{F}_{\theta}$ denotes the SSVEPformer decoder with parameters $\theta$ and $\mathbf{p}_{h} \in \mathbb{R}^{N_f}$ is the vector of posterior probabilities over the $N_f$ stimulus classes. In this study, SSVEPformer is used as the deep decoder to learn temporal and spectral representations from the TD-RCA-enhanced features. The decoder is a modular component of KDHNet and can, in principle, be replaced by another compatible neural network after adapting its input layer to the dimensions of the TD-RCA features.

The predicted label is given by
\begin{equation}
	\hat{y}_{h} = \arg\max_{f \in \{1,\ldots,N_f\}} p_{h,f}.
	\label{eq:predict}
\end{equation}

\subsection{Target-Guided LST Alignment}
In the intra-subject setting, the limited target calibration trials are directly used for training, whereas the inter-subject setting first aligns the source-domain trials to the target subject by LST before training. Let $\mathcal{D}_{s}=\bigcup_{v\neq t}\mathcal{D}_{v}$ denote the source domain, where $t$ is the target subject and $v$ indexes the source subjects. For each source subject $v$ and class $f$, a source template and a target calibration template are computed as
\begin{equation}
\mathbf{S}_{v}^{f}=\frac{1}{n_{v}^{f}}\sum_{y_{v,i}=f}\mathbf{X}_{v,i},
\label{eq:src_template}
\end{equation}
\begin{equation}
	\mathbf{T}_{t}^{f}=\frac{1}{n_{t,\mathrm{cal}}^{f}}\sum_{y_{t,i}=f}\mathbf{X}_{t,i},
	\label{eq:tgt_template}
\end{equation}
where $\{\mathbf{X}_{t,i}, y_{t,i}\}\in \mathcal{D}_{t}^{\mathrm{cal}}$, $n_{v}^{f}$ is the number of available source trials of class $f$, and $n_{t,\mathrm{cal}}^{f}$ is the number of target calibration trials of class $f$.

For each source subject, a class-specific LST matrix is estimated by
\begin{equation}
	\mathbf{A}_{v}^{f}
	=
	\arg\min_{\mathbf{A}}
	\left\|
	\mathbf{A}\mathbf{S}_{v}^{f}-\mathbf{T}_{t}^{f}
	\right\|_{F}^{2},
\end{equation}
whose closed-form solution is
\begin{equation}
	\mathbf{A}_{v}^{f}
	=
	\mathbf{T}_{t}^{f}
	\left(\mathbf{S}_{v}^{f}\right)^{\dagger},
	\label{eq:lst_solution}
\end{equation}
where $(\cdot)^{\dagger}$ denotes the Moore-Penrose pseudoinverse.

Each source-domain trial of class $f$ is then aligned by
\begin{equation}
	\widetilde{\mathbf{X}}_{v,h}^{f}
	=
	\mathbf{A}_{v}^{f}\mathbf{X}_{v,h}^{f}.
	\label{eq:lst_align}
\end{equation}
After LST, all aligned source-domain trials are concatenated with the target calibration trials to form the training set in the inter-subject scenario:
\begin{equation}
	\mathcal{D}_{\mathrm{train}}^{\mathrm{inter}}
	=
	\widetilde{\mathcal{D}}_{s}\cup\mathcal{D}_{t}^{\mathrm{cal}}.
	\label{eq:inter_train}
\end{equation}
This step reduces inter-subject distribution mismatch while retaining subject-specific information provided by the limited target calibration data.

\subsection{Training Objective}

The model is trained using the cross-entropy loss
\begin{equation}
	\mathcal{L}_{\mathrm{CE}}
	=
	-\frac{1}{B}\sum_{i=1}^{B}\sum_{f=1}^{N_f}
	y_{i,f}\log p_{i,f},
	\label{eq:ce}
\end{equation}
where $B$ denotes the batch size, $y_{i,f}$ is the one-hot label of class $f$ for the $i$-th sample, and $p_{i,f}$ is the corresponding predicted probability.

The overall training and inference procedure of KDHNet is summarized in Algorithm~\ref{alg:kdhnet_training}. The TDCA projections are estimated once in closed form and kept fixed during end-to-end training, whereas the TRCA filters and sub-band weights are initialized from their analytical solutions and then updated together with the decoder.

\begin{algorithm}[!t]
	\caption{KDHNet}
	\label{alg:kdhnet_training}
	\begin{algorithmic}[1]
		\REQUIRE Target calibration set $\mathcal{D}_{t}^{\mathrm{cal}}$; target test set $\mathcal{D}_{t}^{\mathrm{te}}$; source domain $\mathcal{D}_{s}$ (inter-subject scenario only); filter bank $\{\mathcal{B}_m\}_{m=1}^{N_m}$; TDCA projections $\{\mathbf{W}_{\mathrm{TD}}^{(m)}\}$; batch normalization and PReLU parameters $\psi$; TRCA filters $\{\mathbf{w}_{\mathrm{TR},f}^{(m)}\}$; sub-band weights $\{\alpha_m\}$; SSVEPformer decoder $\mathcal{F}_{\theta}$; hyperparameters $\epsilon$, $a$, $b$, number of epochs $E$, batch size $B$, learning rate $\eta$
		\STATE \textbf{\# Stage I: Training set construction}
		\IF{inter-subject scenario}
		\FOR{each source subject $v$ and class $f$}
		\STATE Compute $\mathbf{S}_{v}^{f}$ and $\mathbf{T}_{t}^{f}$ by (\ref{eq:src_template}) and (\ref{eq:tgt_template})
		\STATE Estimate $\mathbf{A}_{v}^{f}$ by (\ref{eq:lst_solution}) and align the source trials of class $f$ by (\ref{eq:lst_align})
		\ENDFOR
		\STATE Set $\mathcal{D}_{\mathrm{train}}$ by (\ref{eq:inter_train})
		\ELSE
		\STATE $\mathcal{D}_{\mathrm{train}}\leftarrow\mathcal{D}_{t}^{\mathrm{cal}}$
		\ENDIF
		\STATE Decompose each trial in $\mathcal{D}_{\mathrm{train}}$ into $N_m$ sub-bands by (\ref{eq:filterbank})
		\STATE \textbf{\# Stage II: Analytical initialization of the front-end}
		\FOR{$m=1$ to $N_m$}
		\STATE Estimate $\mathbf{W}_{\mathrm{TD}}^{(m)}$ by (\ref{eq:tdca})
		\STATE Compute $\widehat{\mathbf{Z}}_{f,h}^{(m)}$ by (\ref{eq:bn_prelu})
		\STATE Estimate $\mathbf{w}_{\mathrm{TR},f}^{(m)}$, $f=1,\ldots,N_f$, by (\ref{eq:trca})
		\ENDFOR
		\STATE Initialize $\{\alpha_m\}$ by (\ref{eq:alpha_init}) and randomly initialize $\theta$
		\STATE Freeze $\{\mathbf{W}_{\mathrm{TD}}^{(m)}\}_{m=1}^{N_m}$ (excluded from gradient updates)
		\STATE \textbf{\# Stage III: End-to-end training}
		\FOR{$e=1$ to $E$}
		\FOR{each mini-batch $\mathcal{B}\subset\mathcal{D}_{\mathrm{train}}$ of size $B$}
		\STATE Compute $\mathbf{R}_{h}^{(m)}$ by (\ref{eq:trca_response}) and (\ref{eq:stack})
		\STATE Compute $\widetilde{\mathbf{R}}_{h}$ by (\ref{eq:weighting}) and (\ref{eq:fused})
		\STATE Compute $\mathbf{p}_{h}$ by (\ref{eq:decoder})
		\STATE Compute $\mathcal{L}_{\mathrm{CE}}$ by (\ref{eq:ce})
		\STATE Update $(\psi,\{\mathbf{w}_{\mathrm{TR},f}^{(m)}\},\{\alpha_m\},\theta)$ using $\nabla\mathcal{L}_{\mathrm{CE}}$ with learning rate $\eta$
		\ENDFOR
		\ENDFOR
		\STATE \textbf{\# Inference}
		\FOR{each test trial $\mathbf{X}\in\mathcal{D}_{t}^{\mathrm{te}}$}
		\STATE Compute $\mathbf{p}$ with dropout disabled and predict $\hat{y}$ by (\ref{eq:predict})
		\ENDFOR
		\RETURN Trained KDHNet $(\{\mathbf{W}_{\mathrm{TD}}^{(m)}\},\{\mathbf{w}_{\mathrm{TR},f}^{(m)}\},\mathcal{F}_{\theta})$
	\end{algorithmic}
\end{algorithm}

\section{Experiments}
\label{sec:experiments}

\subsection{Datasets}

Three publicly available SSVEP datasets were used in this study, as summarized in Table~\ref{tab:dataset_summary}.
\begin{enumerate}
	\item Dataset~I (12-target)~\cite{Nakanishi2015} contains recordings from 10 subjects and 12 targets, with 15 trials per class and 8 occipital and parieto-occipital channels.
	\item Dataset~II (Benchmark)~\cite{BenchmarkDataset} contains recordings from 35 subjects and 40 targets, with 6 trials per class. Following common practice, 9 posterior channels were used.
	\item Dataset~III (BETA)~\cite{BetaDataset} contains recordings from 70 subjects under the same 40-target paradigm, with 4 trials per class and the same 9 posterior channels.
\end{enumerate}

\begin{table}[t]
	\centering
	\setlength{\tabcolsep}{1mm}
	\renewcommand\arraystretch{1}
	\caption{Main characteristics of the three SSVEP datasets. $f_s$ denotes the sampling rate.}
	\label{tab:dataset_summary}
	\begin{tabular}{l|ccccc}
		\toprule
		Dataset & \# Subject & \# Classes & \# Channels & $f_s$ (Hz) & \# Trials/class \\
		\midrule
		I (12-target) & 10 & 12 & 8 & 256 & 15 \\
		II (Benchmark) & 35 & 40 & 9 & 250 & 6 \\
		III (BETA) & 70 & 40 & 9 & 250 & 4 \\
		\bottomrule
	\end{tabular}
\end{table}

\subsection{Preprocessing and Experimental Protocol}

All datasets were evaluated with a fixed 1.0~s window in the split-ratio experiments and with multiple windows from 0.4 s to 1.2 s in the time-window experiments. Filter-bank preprocessing was applied before decoding.

Two limited-calibration protocols were considered. In the intra-subject setting, the training and test data were drawn from the same subject. For a class with $K$ trials and training ratio $r$, the number of training trials was $n_{\mathrm{tr}} = \lfloor rK \rfloor$, and the remaining trials were used for testing. We evaluated $r \in \{0.2, 0.4, 0.6, 0.8\}$ on Dataset~I with five folds, $r \in \{0.35, 0.5, 0.7, 0.85\}$ on Dataset~II with six folds, and $r \in \{0.5, 0.75\}$ on Dataset~III with four folds. Each fold used a circular contiguous split within each class, with a step size of 3 trials for Dataset~I and 1 trial for Datasets~II and III.

In the inter-subject setting, LOSO evaluation was adopted. For each target subject, the remaining subjects were treated as the source domain, while one calibration trial per class from the target subject was reserved to estimate class-wise LST matrices. The aligned source-domain trials and the target calibration trials were then used for training, and the remaining target-subject trials were used for testing.

\subsection{Compared Approaches}

To validate the effectiveness of KDHNet, we compared it with representative template-based algorithms and deep learning decoders. All approaches received the same filter-bank input with $N_m=3$ sub-bands, and the template-based algorithms combined the scores of the sub-bands with the weights $m^{-1.25}+0.25$.
\begin{enumerate}
	\item FBCCA \cite{FBCCA} computes, in each sub-band, the canonical correlation between the test trial and the reference signal of each class, and selects the class with the largest weighted sum of squared correlations. In this study, the class-mean training trials of each sub-band were used as the reference signals, so FBCCA used the same calibration data as the other approaches.
	\item TRCA \cite{TRCA} learns, in each sub-band, a spatial filter for each class that maximizes the reproducibility of the filtered signal across training trials. The ensemble version was used, in which the filters of all classes are applied to both the test trial and the class templates, and the class with the largest weighted correlation is selected.
	\item TDCA \cite{Liu2021TDCA} learns a spatial projection that maximizes the ratio of between-class to within-class scatter. It was implemented with the same discriminant projection as the TDCA stage of KDHNet, using one component per sub-band, and a test trial was assigned to the class whose projected template was most correlated with the projected trial.
	\item EEGNet \cite{EEGNet} is a compact CNN that applies a temporal convolution, a depthwise spatial convolution, and a separable convolution to the time-domain EEG.
	\item CCNN \cite{CCNN} is a CNN that takes as input the complex spectrum of each channel, namely the concatenated real and imaginary parts of its fast Fourier transform.
	\item FBtCNN \cite{Ding2021FBtCNN} decomposes the time-domain input into several sub-bands, processes each sub-band with a shared temporal CNN, and fuses the sub-band features for classification.
	\item SSVEPNet \cite{SSVEPNet} combines a CNN with a bidirectional long short-term memory network, and uses spectral normalization and label smoothing to reduce overfitting to limited training data.
	\item SSVEPformer \cite{SSVEPformer} applies a convolution-based Transformer encoder to the complex spectrum of the input. It shares its architecture with the decoder of KDHNet, so the comparison between the two reflects the contribution of the TD-RCA front-end.
\end{enumerate}

\subsection{Implementation Details}

\begin{table}[t]
	\centering
	\setlength{\tabcolsep}{2mm}
	\renewcommand\arraystretch{1}
	\caption{Key architectural settings of KDHNet. $f_{\mathrm{start}}$ denotes the lowest stimulus frequency of each dataset.}
	\label{tab:kdhnet_arch}
	\begin{tabular}{l|l}
		\toprule
		Item & Setting \\
		\midrule
		Input representation & Multiple sub-band EEG trials \\
		Number of sub-bands $N_m$ & 3 \\
		TDCA components & 3 for intra-subject, 1 for inter-subject \\
		TRCA components per class & 1 \\
		Dropout before TRCA & 0.5 \\
		Sub-band weighting & Learnable, $\alpha_m^{(0)} = m^{-1.25}+0.25$ \\
		Decoder & SSVEPformer \\
		Transformer depth & 2 \\
		Spectral convolution kernel length & 31 \\
		Decoder dropout & 0.5 \\
		FFT range & $[f_{\mathrm{start}}, 64]$ Hz \\
		Loss function & Cross-entropy \\
		\bottomrule
	\end{tabular}
\end{table}

All experiments used a unified PyTorch framework (\href{https://github.com/2665023991/KDHNet}{KDHNet code}). The architecture of KDHNet is summarized in Table~\ref{tab:kdhnet_arch}. KDHNet was trained for $100$ epochs with batch size $30$ using stochastic gradient descent with learning rate $10^{-3}$, momentum 0.9, weight decay $10^{-4}$, and cosine-annealing decay. Each baseline used its recommended training configuration. The deep baselines took the $N_mN_c$-channel filter-bank signal as input. EEGNet, CCNN, FBtCNN, and SSVEPNet were trained with Adam for 300, 300, 500, and 400 epochs, respectively, with a learning rate of $10^{-3}$ except for EEGNet ($5\times10^{-3}$). SSVEPformer was trained with the same optimizer, learning rate, and number of epochs as KDHNet.

\subsection{Evaluation Metrics and Statistical Analysis}

The decoding performance was evaluated using classification accuracy and ITR. For a test set with $N$ trials, the classification accuracy is $\mathrm{ACC} = N_{\mathrm{correct}}/N$, where $N_{\mathrm{correct}}$ denotes the number of correctly classified trials. The ITR, measured in bits/min, was computed as
\begin{equation}
	\begin{aligned}
		\mathrm{ITR}
		=&\frac{60}{T_{\mathrm{d}}}\Bigg[
		\log_2 N_f
		+ P \log_2 P \\
		&\quad
		+ (1-P)\log_2 \left(\frac{1-P}{N_f-1}\right)
		\Bigg],
	\end{aligned}
\end{equation}
where $N_f$ is the number of stimulus classes, $P$ is the classification accuracy, and $T_{\mathrm{d}}$ is the time in seconds required for one decision. In this study, $T_{\mathrm{d}}$ equals the data window length used for decoding.

For both protocols, the results were first averaged across folds within each subject and then reported as the mean and standard deviation across subjects. KDHNet was compared with each baseline using two-sided Wilcoxon signed-rank tests \cite{Wilcoxon1945} on the subject-wise accuracies. The samples were paired over subjects, giving 10, 35, and 70 pairs on Datasets~I, II, and III, respectively. The $p$-values were adjusted with the Holm-Bonferroni procedure \cite{Holm1979} within each dataset and protocol, that is, over the eight baselines compared with KDHNet. With 10 pairs, the exact two-sided test cannot return a $p$-value below $1.95\times10^{-3}$, which becomes $1.56\times10^{-2}$ after the correction over eight baselines, so a comparison on Dataset~I carries a single asterisk even when KDHNet outperforms the baseline for every subject.

\section{Results}
\subsection{Intra-Subject Limited-Calibration Results}
\label{sec:intra}

\begin{figure*}[t]
	\centering
	\includegraphics[width=\textwidth]{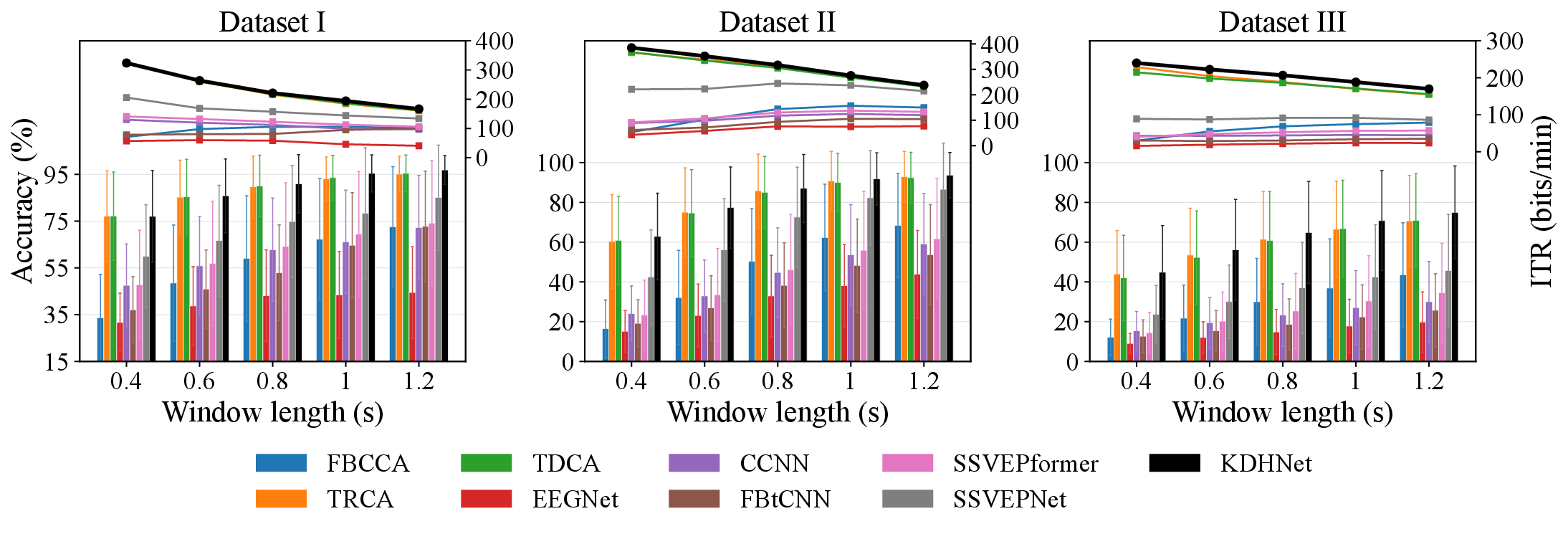}
	\caption{Intra-subject decoding accuracy (bars, left axis) and ITR (lines, right axis) versus the analysis window length for the nine methods on Datasets~I, II, and III, with one panel per dataset. Error bars denote one standard deviation across subjects. The training ratio was fixed at 0.2 for Dataset~I and 0.5 for Datasets~II and III. KDHNet is drawn in black. The full per-window results are provided in Table~S1 of the supplementary material.}
	\label{fig:curve_window}
\end{figure*}

We first evaluated intra-subject decoding with five analysis windows from 0.4 to 1.2~s. The training ratio was fixed at 0.2 for Dataset~I and 0.5 for Datasets~II and III. Fig.~\ref{fig:curve_window} shows the results, and the full per-window results are given in Table~S1 of the supplementary material. Observe that:
\begin{enumerate}
	\item KDHNet achieved the highest mean accuracy and ITR in 14 of the 15 window-dataset settings. The only exception was the 0.4~s window on Dataset~I, where TDCA exceeded KDHNet by 0.10\% in accuracy and TRCA exceeded it by 0.51~bits/min in ITR.
	\item Averaged over the five windows, KDHNet outperformed the strongest template-based baseline by 0.90\% on Dataset~I (TDCA), 1.60\% on Dataset~II (TRCA), and 3.12\% on Dataset~III (TRCA). The margin increased from the 12-class Dataset~I to the low-SNR Dataset~III.
	\item The effect of the window length on this margin was dataset-dependent. From 0.4 to 1.2~s, the margin increased from $-0.10\%$ to 1.46\% on Dataset~I and from 0.87\% to 3.96\% on Dataset~III, but decreased from 2.02\% to 0.84\% on Dataset~II.
	\item Deep baselines without an analytical front-end performed substantially worse. Averaged over the five windows, SSVEPformer, which shares its architecture with the decoder of KDHNet, was lower than KDHNet by 26.76\%, 38.44\%, and 37.35\% on Datasets~I, II, and III, respectively.
\end{enumerate}
Overall, the analytical front-end accounted for most of the accuracy gain over the deep baselines under limited calibration, and the learnable decoder provided a further improvement over the template-based algorithms in most settings.

\subsection{Effect of Training Ratio}
\label{sec:ratio}

\begin{figure*}[t]
	\centering
	\includegraphics[width=.9\textwidth]{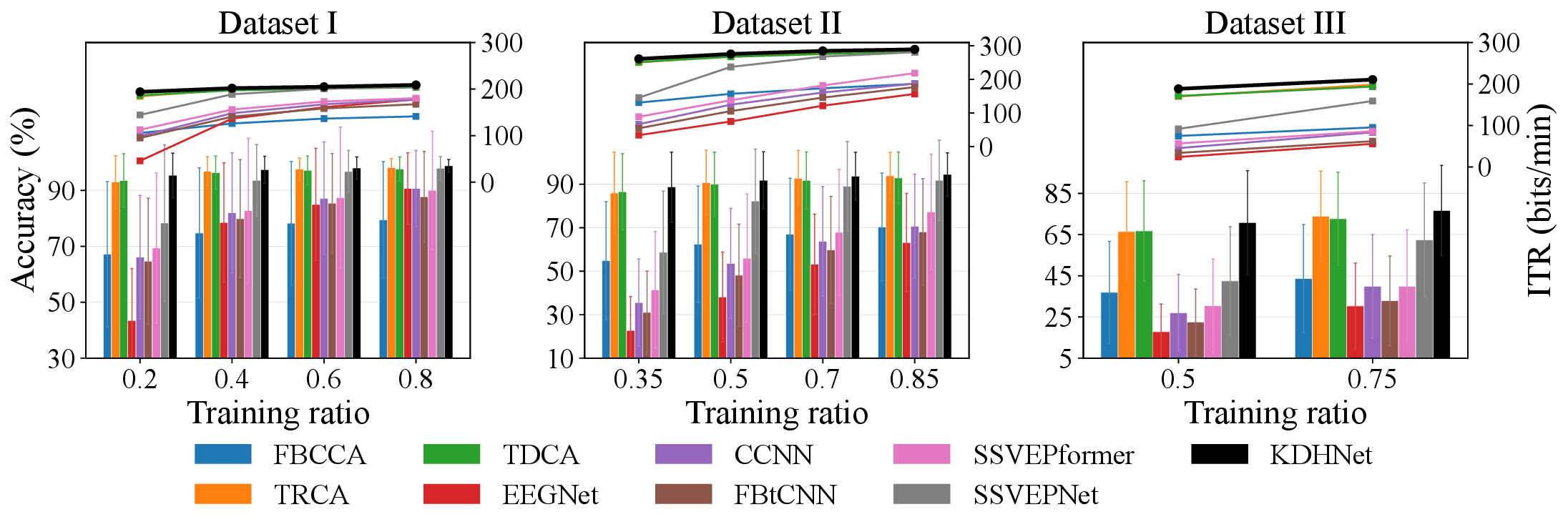}
	\caption{Intra-subject decoding accuracy (bars, left axis) and ITR (lines, right axis) at the 1.0~s window versus the training ratio for the nine methods on Datasets~I, II, and III, with one panel per dataset. Error bars denote one standard deviation across subjects. KDHNet is drawn in black. The full results are provided in Table~S2 of the supplementary material.}
	\label{fig:curve_ratio}
\end{figure*}

We next varied the amount of calibration data with a fixed 1.0~s window. Fig.~\ref{fig:curve_ratio} shows the results, and the full results are given in Table~S2 of the supplementary material. Observe that:
\begin{enumerate}
	\item KDHNet achieved the highest mean accuracy and ITR in all ten ratio-dataset settings.
	\item The margin over the strongest template-based baseline was largest at the smallest training ratio, reaching 1.86\%, 2.14\%, and 4.03\% on Datasets~I, II, and III, respectively. At the largest training ratio, it decreased to 0.69\%, 0.60\%, and 2.74\%.
	\item The deep baselines were most affected by the reduction of calibration data. At $r=0.35$ on Dataset~II, SSVEPformer achieved an accuracy of 41.40\%, which was 47.25\% lower than that of KDHNet.
\end{enumerate}
Overall, the advantage of KDHNet was concentrated in the calibration-scarce regime targeted by its design.

\subsection{Per-Subject Improvement}

\begin{figure*}[t]
	\centering
	\includegraphics[width=0.25\textwidth]{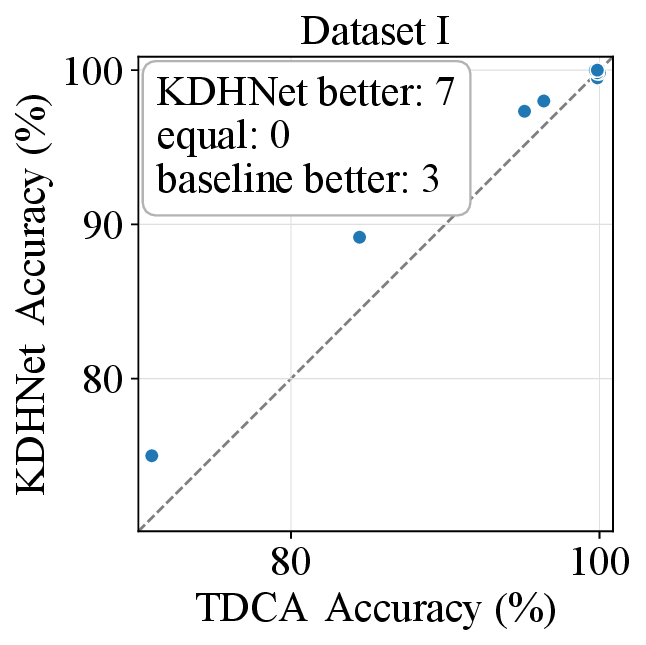}
	\includegraphics[width=0.25\textwidth]{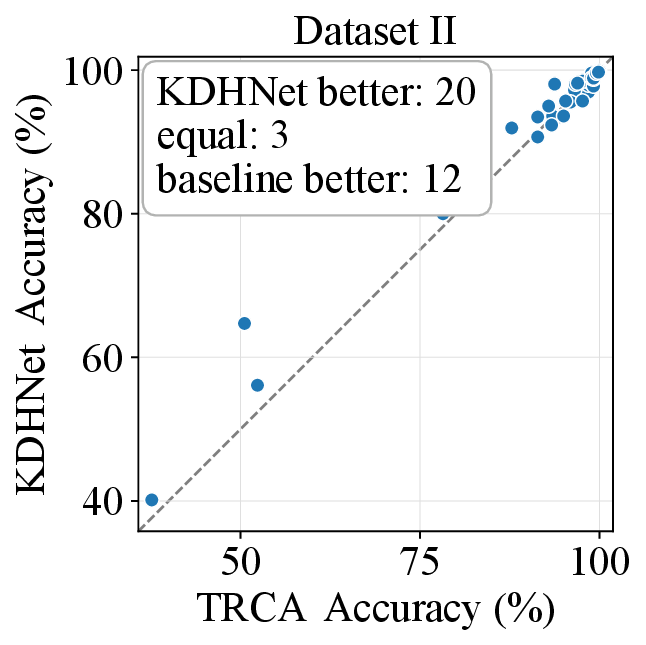}
	\includegraphics[width=0.25\textwidth]{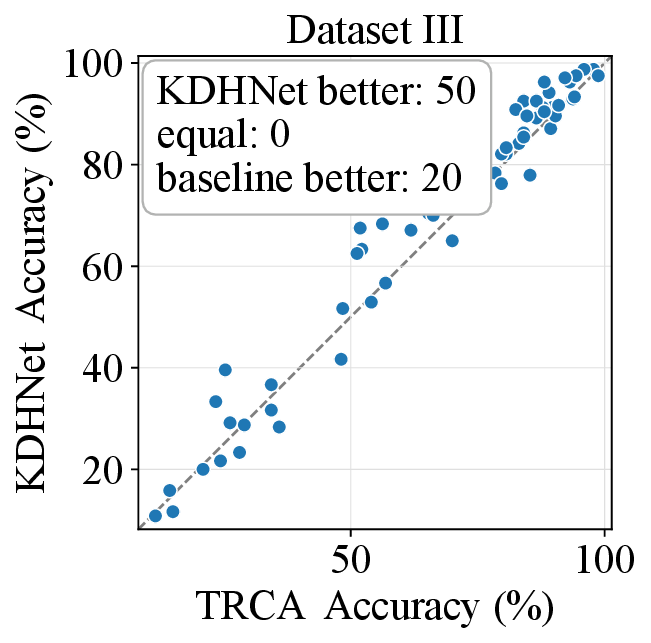}
	\caption{Per-subject accuracy of KDHNet ($y$-axis) against the strongest template-based baseline ($x$-axis) at the 1.0~s intra-subject setting: (left) Dataset~I versus TDCA, (middle) Dataset~II versus TRCA, and (right) Dataset~III versus TRCA. Each marker denotes one subject. Markers above the $y=x$ line indicate subjects for whom KDHNet outperformed the baseline.}
	\label{fig:per_subject_scatter}
\end{figure*}

The per-subject accuracies of KDHNet and the strongest template-based baseline at the 1.0~s intra-subject setting are compared in Fig.~\ref{fig:per_subject_scatter}, using TDCA on Dataset~I and TRCA on Datasets~II and III. KDHNet outperformed the baseline for 7 of 10 subjects on Dataset~I, 20 of 35 subjects on Dataset~II, and 50 of 70 subjects on Dataset~III. On Datasets~I and II, the improvement was concentrated in the lowest quartile of baseline accuracy, where the mean gains were 4.98\% and 4.03\%, compared with 0.52\% and 0.05\% for the remaining subjects. On Dataset~III, the improvement was distributed more evenly, with mean gains of 2.99\% in the lowest quartile and 4.83\% for the remaining subjects. These results indicate that, on Datasets~I and II, the mean improvement mainly came from subjects who were poorly decoded by the template-based algorithms, whereas on Dataset~III it was spread across the accuracy range.

\subsection{Inter-Subject Results}

For inter-subject decoding, we used LOSO evaluation with a 1.0~s window, one calibration trial per class, and class-wise LST alignment of the source data. Table~\ref{tab:main_cross_1s} reports the results. Observe that:
\begin{enumerate}
	\item KDHNet achieved the highest accuracy and ITR on Datasets~I and II. On Dataset~I, it exceeded the second-best method, SSVEPformer, by 1.93\% in accuracy. On Dataset~II, it exceeded SSVEPformer by 3.28\% and TRCA by 17.00\%.
	\item The template-based algorithms were less effective on the 40-class Dataset~II. TRCA and TDCA achieved 57.44\% and 56.92\%, lower than CCNN (61.83\%) and SSVEPformer (71.16\%). This is consistent with the difficulty of estimating reliable class templates from a single target trial per class when the number of classes is large.
	\item On Dataset~III, TRCA achieved the highest accuracy (46.29\%), and KDHNet ranked second (43.31\%). This difference of 2.98\% was not statistically significant (Section~\ref{sec:significance}).
\end{enumerate}
Overall, KDHNet achieved the best or second-best performance on all three datasets. The residual gap on Dataset~III suggests that one target trial per class is insufficient for the deep decoder to adapt to the heterogeneous, low-SNR recordings of BETA, which motivates richer target-domain adaptation (Section~\ref{sec:discussion}).

\begin{table*}[t]
\centering
\setlength{\tabcolsep}{2mm}
\renewcommand\arraystretch{1}
\caption{Inter-subject accuracy (\%) and ITR (bits/min) at the 1.0~s window under LOSO evaluation, with one target calibration trial per class and class-wise LST alignment. Values are the mean$\pm$standard deviation across subjects. The best result in each column is marked in bold, and the second-best by an underline.}
\label{tab:main_cross_1s}
\begin{tabular}{l|cc|cc|cc}
\toprule
\multirow{2.5}{*}{Method} & \multicolumn{2}{c|}{Dataset I} & \multicolumn{2}{c|}{Dataset II} & \multicolumn{2}{c}{Dataset III} \\
\cmidrule{2-7}
 & ACC & ITR & ACC & ITR & ACC & ITR \\
\midrule
FBCCA & 64.26$\pm$21.56 & 93.02$\pm$54.41 & 58.51$\pm$22.99 & 139.13$\pm$77.72 & 39.47$\pm$22.30 & 79.01$\pm$65.96 \\
TRCA & 88.56$\pm$13.85 & 167.72$\pm$47.50 & 57.44$\pm$28.99 & 141.26$\pm$97.68 & \textbf{46.29$\pm$21.73} & \textbf{97.98$\pm$67.02} \\
TDCA & 87.64$\pm$13.78 & 164.15$\pm$46.82 & 56.92$\pm$29.07 & 139.51$\pm$97.23 & 42.49$\pm$20.96 & 86.18$\pm$62.13 \\
EEGNet & 73.68$\pm$21.45 & 120.10$\pm$58.48 & 43.37$\pm$20.28 & 88.70$\pm$59.52 & 22.61$\pm$13.38 & 32.22$\pm$31.23 \\
CCNN & 83.63$\pm$18.64 & 152.00$\pm$57.89 & 61.83$\pm$23.96 & 152.36$\pm$79.50 & 35.07$\pm$20.39 & 65.73$\pm$57.51 \\
FBtCNN & 71.37$\pm$20.61 & 112.15$\pm$52.86 & 42.94$\pm$20.17 & 87.27$\pm$59.97 & 20.44$\pm$12.31 & 27.27$\pm$27.48 \\
SSVEPformer & \underline{88.92$\pm$15.70} & \underline{171.19$\pm$53.84} & \underline{71.16$\pm$26.60} & \underline{191.68$\pm$92.42} & 37.67$\pm$22.06 & 74.34$\pm$63.91 \\
SSVEPNet & 74.56$\pm$18.22 & 120.61$\pm$49.95 & 31.70$\pm$21.87 & 60.63$\pm$57.71 & 9.02$\pm$8.22 & 7.73$\pm$16.90 \\
KDHNet & \textbf{90.85$\pm$11.87} & \textbf{175.84$\pm$43.37} & \textbf{74.44$\pm$27.37} & \textbf{207.15$\pm$95.28} & \underline{43.31$\pm$27.30} & \underline{95.50$\pm$83.37} \\
\bottomrule
\end{tabular}
\end{table*}

\subsection{Statistical Significance}
\label{sec:significance}

Table~\ref{tab:significance} reports the Holm-corrected $p$-values of the Wilcoxon signed-rank tests of KDHNet against each baseline on the subject-wise accuracies at the 1.0~s window. All significant differences favored KDHNet. Observe that:
\begin{enumerate}
	\item KDHNet significantly outperformed all five deep baselines in all six dataset-protocol settings, except SSVEPformer under the inter-subject protocol on Dataset~I ($p=0.625$).
	\item KDHNet significantly outperformed FBCCA in five of the six settings and TDCA in three of the six settings. The exceptions for TDCA were intra-subject Dataset~I ($p=0.129$), where both methods exceeded 93\% accuracy, inter-subject Dataset~I ($p=0.082$), and inter-subject Dataset~III ($p=0.808$).
	\item Against TRCA, the improvement was significant on intra-subject Dataset~III and inter-subject Dataset~II. The differences in the other four settings were not significant, including inter-subject Dataset~III, where TRCA was numerically better ($p=0.288$).
\end{enumerate}
Overall, KDHNet was significantly better than the deep baselines in 29 of the 30 comparisons, and it was significantly better than or statistically comparable to the template-based algorithms in all settings.

\begin{table}[t]
\centering
\setlength{\tabcolsep}{2mm}
\renewcommand\arraystretch{1}
\caption{Two-sided Wilcoxon signed-rank $p$-values of KDHNet against each baseline, computed on the subject-wise accuracies at the 1.0~s window and adjusted with the Holm-Bonferroni procedure over the eight baselines in each column. All significant differences favor KDHNet. Statistical significance is denoted by * ($p<0.05$), ** ($p<0.005$), and *** ($p<0.001$).}
\label{tab:significance}
\begin{tabular}{l|ccc}
\toprule
Baseline & Dataset I & Dataset II & Dataset III \\
\midrule
\multicolumn{4}{c}{Intra-subject} \\
\midrule
FBCCA & 0.016* & $<$0.001*** & $<$0.001*** \\
TRCA & 0.129 & 0.087 & $<$0.001*** \\
TDCA & 0.129 & $<$0.001*** & $<$0.001*** \\
EEGNet & 0.016* & $<$0.001*** & $<$0.001*** \\
CCNN & 0.016* & $<$0.001*** & $<$0.001*** \\
FBtCNN & 0.016* & $<$0.001*** & $<$0.001*** \\
SSVEPformer & 0.016* & $<$0.001*** & $<$0.001*** \\
SSVEPNet & 0.016* & $<$0.001*** & $<$0.001*** \\
\midrule
\multicolumn{4}{c}{Inter-subject} \\
\midrule
FBCCA & 0.016* & $<$0.001*** & 0.288 \\
TRCA & 0.168 & 0.005* & 0.288 \\
TDCA & 0.082 & 0.005* & 0.808 \\
EEGNet & 0.016* & $<$0.001*** & $<$0.001*** \\
CCNN & 0.016* & $<$0.001*** & $<$0.001*** \\
FBtCNN & 0.016* & $<$0.001*** & $<$0.001*** \\
SSVEPformer & 0.625 & 0.002** & $<$0.001*** \\
SSVEPNet & 0.016* & $<$0.001*** & $<$0.001*** \\
\bottomrule
\end{tabular}
\end{table}

\section{Ablation Study}
\label{sec:ablation}

\subsection{Component-Wise Ablation}

We ablated the three knowledge layers of KDHNet, namely the TDCA stage, the TRCA stage, and the sub-band weighting layer. Each layer was run in one of four modes: analytical-frozen, analytical-trainable, random-trainable, or disabled. The default KDHNet keeps the TDCA stage analytical-frozen and the other two layers analytical-trainable. Each row of Table~\ref{tab:ablation_init_modes} changes exactly one layer while the other two keep their default modes. All runs used the intra-subject protocol with the 1.0~s window on the three datasets.

\begin{table*}[t]
	\centering
	\setlength{\tabcolsep}{2mm}
	\renewcommand\arraystretch{1}
	\caption{Component-wise ablation of KDHNet on the three datasets under the intra-subject protocol with the 1.0~s window. The default configuration keeps the TDCA stage analytical-frozen and the TRCA stage and sub-band weighting layer analytical-trainable. Each block modifies exactly one component relative to this default. Values are the mean$\pm$standard deviation across subjects, and Mean denotes the average over the three datasets. The best result in each column is marked in bold, and the second-best by an underline.}
	\label{tab:ablation_init_modes}
	\begin{tabular}{l|cccc|cccc}
		\toprule
		\multirow{2.5}{*}{Ablation}
		& \multicolumn{4}{c|}{ACC (\%)}
		& \multicolumn{4}{c}{ITR (bits/min)} \\
		\cmidrule{2-9}
		& I & II & III & Mean
		& I & II & III & Mean \\
		\midrule
		Default (KDHNet)
		& \textbf{95.42$\pm$7.92} & 91.83$\pm$13.10 & \textbf{70.82$\pm$25.14} & \textbf{86.02}
		& \textbf{194.02$\pm$32.04} & 275.67$\pm$56.85 & \textbf{188.35$\pm$89.72} & \textbf{219.35} \\
		\midrule
		\multicolumn{9}{c}{(a) TDCA stage (default: analytical-frozen)} \\
		\midrule
		Analytical-trainable
		& 64.78$\pm$25.61 & 37.03$\pm$24.11 & 24.07$\pm$17.63 & 41.96
		& 99.07$\pm$63.78 & 75.24$\pm$71.75 & 39.04$\pm$43.78 & 71.12 \\
		Random-trainable
		& 79.53$\pm$22.16 & 75.86$\pm$25.16 & 43.52$\pm$26.70 & 66.31
		& 142.18$\pm$61.85 & 211.65$\pm$90.03 & 96.61$\pm$82.98 & 150.15 \\
		Disabled
		& \underline{95.18$\pm$8.25} & \textbf{92.09$\pm$12.93} & 70.17$\pm$25.39 & 85.81
		& \underline{192.95$\pm$32.87} & \underline{277.06$\pm$56.58} & 186.14$\pm$90.77 & 218.72 \\
		\midrule
		\multicolumn{9}{c}{(b) TRCA stage (default: analytical-trainable)} \\
		\midrule
		Analytical-frozen
		& \textbf{95.42$\pm$7.91} & 91.73$\pm$13.28 & \underline{70.63$\pm$25.12} & \underline{85.93}
		& \textbf{194.02$\pm$32.00} & 275.28$\pm$57.44 & 187.50$\pm$89.99 & \underline{218.93} \\
		Random-trainable
		& 93.32$\pm$11.53 & 91.96$\pm$14.01 & 67.84$\pm$26.18 & 84.37
		& 187.15$\pm$41.55 & \textbf{277.22$\pm$59.84} & 177.60$\pm$92.93 & 213.99 \\
		Disabled
		& 93.02$\pm$11.99 & 90.77$\pm$15.52 & 64.76$\pm$27.10 & 82.85
		& 185.79$\pm$43.79 & 271.88$\pm$64.01 & 167.09$\pm$93.53 & 208.25 \\
		\midrule
		\multicolumn{9}{c}{(c) Sub-band weighting layer (default: analytical-trainable)} \\
		\midrule
		Analytical-frozen
		& 94.93$\pm$8.68 & 91.75$\pm$13.26 & \underline{70.63$\pm$25.13} & 85.77
		& 192.12$\pm$34.48 & 275.38$\pm$57.36 & \underline{187.53$\pm$90.01} & 218.34 \\
		Random-trainable
		& 94.27$\pm$9.72 & 91.75$\pm$13.53 & 70.06$\pm$25.11 & 85.36
		& 189.78$\pm$36.90 & 275.57$\pm$57.84 & 185.19$\pm$89.75 & 216.84 \\
		Disabled
		& 94.93$\pm$8.68 & \underline{92.02$\pm$13.20} & 70.21$\pm$24.64 & 85.72
		& 192.12$\pm$34.48 & 276.81$\pm$56.94 & 185.28$\pm$88.44 & 218.07 \\
		\bottomrule
	\end{tabular}
\end{table*}

Table~\ref{tab:ablation_init_modes} reports the results. Observe that:
\begin{enumerate}
	\item The default configuration achieved the highest three-dataset mean accuracy (86.02\%) and ITR (219.35~bits/min). No single-component modification improved either mean, although some variants were marginally better on individual datasets, for example disabling the TDCA stage on Dataset~II (92.09\% versus 91.83\%).
	\item The TDCA stage performed well only when it was not trained. Keeping its analytical solution frozen and disabling it gave mean accuracies of 86.02\% and 85.81\%, respectively. Training it from the analytical solution reduced the mean accuracy to 41.96\%, and training it from random initialization reached only 66.31\%. The accuracy drop of the analytical-trainable variant was 30.64\% on the 12-class Dataset~I, and 54.80\% and 46.75\% on the 40-class Datasets~II and III.
	\item The TRCA stage and the sub-band weighting layer were much less sensitive to the mode choice. Across their non-default modes, the mean accuracy decreased by at most 3.17\% for the TRCA stage and 0.66\% for the sub-band weighting layer. For both layers, initialization from the closed-form solution followed by end-to-end training gave the best mean accuracy.
\end{enumerate}

These results clarify the role of the TDCA stage. Because disabling it changed the mean accuracy by only 0.21\%, the TRCA path alone can supply the discriminative features. Training it, in contrast, degraded the accuracy substantially on all three datasets. The TDCA stage therefore acts as a structural prior that constrains the input of the trainable layers, rather than as an additional feature stream. Freezing it also improved the consistency across subjects. On Datasets~I and II, the standard deviation of accuracy across subjects was 7.92\% and 13.10\% with the frozen TDCA stage, compared with 22.16\% and 25.16\% for the random-trainable variant. Overall, the default configuration, which freezes the TDCA stage and fine-tunes the TRCA and sub-band weighting layers from their analytical solutions, was the most accurate among the evaluated configurations.

\subsection{Replacing the Deep Decoder}
\label{sec:decoder_swap}

\begin{figure}[t]
	\centering
	\includegraphics[width=\linewidth]{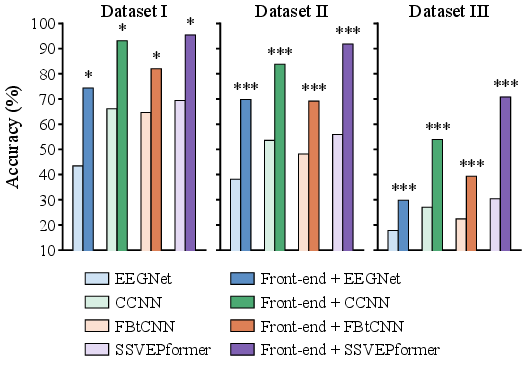}
	\caption{Accuracy of four deep decoders with and without the TD-RCA front-end under the intra-subject protocol with the 1.0~s window ($r=0.2$ on Dataset~I and $r=0.5$ on Datasets~II and III). For each decoder, the light and dark bars of the same color show the mean accuracy across subjects without and with the TD-RCA knowledge front-end (Front-end~+ decoder in the legend), respectively. All models were evaluated at the final training epoch. The SSVEPformer bars without and with the front-end correspond to the SSVEPformer baseline and KDHNet in Section~\ref{sec:intra}, respectively. Statistical significance of the improvement, assessed by two-sided Wilcoxon signed-rank tests over subjects with Holm-Bonferroni correction over the four decoders of each dataset, is denoted by * ($p<0.05$), ** ($p<0.005$), and *** ($p<0.001$).}
	\label{fig:decoder_swap}
\end{figure}

To examine whether the benefit of the TD-RCA front-end depends on the SSVEPformer decoder, we replaced the decoder with three other deep baselines, namely EEGNet, CCNN, and FBtCNN. As described in Section~\ref{sec:methods}, the input layer of each decoder was adapted to the $N_mN_f$ channels of the front-end features. For CCNN, which takes a complex spectrum as input, the spectral transform of its original implementation was applied to the front-end features. The front-end was configured as in KDHNet, and each decoder used its default training configuration. All models were evaluated under the intra-subject protocol with the 1.0~s window, using $r=0.2$ on Dataset~I and $r=0.5$ on Datasets~II and III. The results without the front-end are those of the corresponding deep baselines in Section~\ref{sec:intra}, and the SSVEPformer results with the front-end are those of KDHNet. Fig.~\ref{fig:decoder_swap} shows the results. Observe that:
\begin{enumerate}
	\item The front-end improved the mean accuracy of all four decoders on all three datasets, and all 12 improvements were statistically significant. The gains ranged from 11.97\% (EEGNet on Dataset~III) to 40.41\% (SSVEPformer on Dataset~III).
	\item The improvements were consistent across subjects. On Datasets~I and II, every subject improved with every decoder. On Dataset~III, all 70 subjects improved with CCNN and SSVEPformer, 69 with FBtCNN, and 56 with EEGNet.
	\item With the front-end, SSVEPformer achieved the highest accuracy on all three datasets, which supports its choice as the default decoder of KDHNet. CCNN ranked second, with accuracies of 93.11\%, 83.79\%, and 53.91\% on Datasets~I, II, and III, respectively.
\end{enumerate}
Overall, the TD-RCA front-end improved every evaluated decoder, indicating that its benefit is not tied to a specific architecture.

\section{Interpretability Analysis}

\subsection{Spatial-Filter Structure}
\label{sec:filter_viz}

\begin{figure}[htpb] \centering
\includegraphics[width=.9\linewidth,clip]{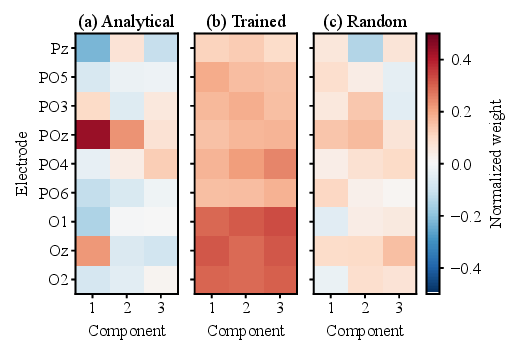}
\caption{Subject-averaged TDCA spatial filter of the first filter-bank sub-band on Dataset~II. The vertical axis indexes the nine occipito-parietal electrodes, and the horizontal axis indexes the three spatial components. (a) Closed-form analytical solution, as used in the frozen KDHNet front-end. (b) The same filter initialized at the closed-form solution and trained end-to-end. (c) A randomly initialized filter trained end-to-end. Each subject's filter was sign-aligned and unit-normalized before averaging.} \label{fig:filter_maps}
\end{figure}

Section~\ref{sec:ablation} showed that the TDCA stage was effective only when frozen, and the learned filter weights provide an explanation. We took the TDCA filter of the first sub-band, sign-aligned and unit-normalized each spatial component, and averaged it over the 35 subjects of Dataset~II under three regimes, namely the closed-form solution, the same solution after end-to-end training, and a randomly initialized filter after end-to-end training. Fig.~\ref{fig:filter_maps} shows the results. The frozen analytical filter concentrated its weights on the medial occipito-parietal electrodes, where SSVEP responses are typically strongest. After end-to-end training, this focal structure disappeared. Both the analytically and randomly initialized filters converged to diffuse maps that were inconsistent across subjects, so their subject average showed no clear spatial pattern. The median cosine similarity between the trained and analytical filters was 0.00, comparable to that between two random filters. On Dataset~II, the corresponding mean accuracies of the three regimes were 91.83\%, 37.03\%, and 75.86\% (Table~\ref{tab:ablation_init_modes}). End-to-end training therefore replaced a focal occipital filter with a spatially unstructured one, which is consistent with the accuracy decrease of the trainable TDCA variants.

\subsection{Feature Separability}
\label{sec:feature_sep}

\begin{figure}[htpb] \centering
\includegraphics[width=.9\linewidth,clip]{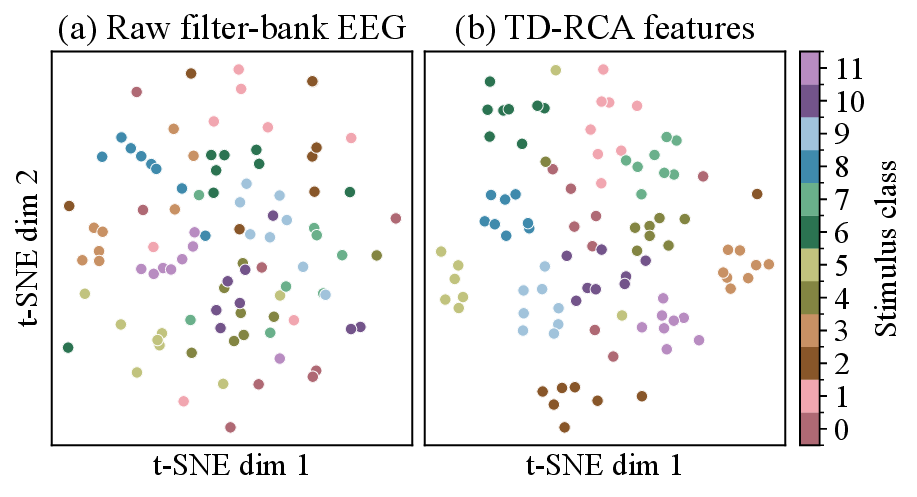}
\caption{Two-dimensional t-SNE embedding of the test trials of a representative Dataset~I subject, colored by stimulus class, for (a) the raw filter-bank EEG and (b) the TD-RCA front-end features fed to the decoder. For this subject, the five-nearest-neighbor accuracy increased from 45.8\% on the raw EEG to 83.4\% on the front-end features.} \label{fig:tsne_features}
\end{figure}

The front-end supplies the decoder with the stacked, sub-band-weighted TRCA responses $\widetilde{\mathbf{R}}_{h}$ instead of raw EEG. Fig.~\ref{fig:tsne_features} shows a two-dimensional t-SNE embedding of the test trials of one Dataset~I subject, colored by stimulus class. In the raw filter-bank signal, trials of different classes overlapped heavily, whereas the front-end features formed compact class-consistent clusters. A five-nearest-neighbor classifier was used to quantify this difference. Averaged over the ten subjects of Dataset~I, its accuracy increased from 53.5\% on the raw signal to 91.6\% on the front-end features. This gain was obtained from the analytical stages alone, mainly the per-class TRCA filters, without any decoder training. The decoder therefore receives an input that is already largely class-separable, and it mainly needs to refine the decision boundary instead of learning discriminative structure from few calibration trials.

\section{Discussion}
\label{sec:discussion}

The advantage of KDHNet was largest when calibration data were scarce. At the smallest training ratio, its margin over the strongest template-based baseline reached 1.86\%, 2.14\%, and 4.03\% on Datasets~I, II, and III, respectively (Section~\ref{sec:ratio}). The main exception was inter-subject decoding on Dataset~III, where TRCA exceeded KDHNet by 2.98\%, although the difference was not significant. With one calibration trial per class, TRCA can still construct usable templates, whereas the deep decoder must generalize across the heterogeneous, low-SNR BETA recordings with very little target-subject supervision.

The ablation and interpretability analyses lead to a consistent interpretation. Removing the TDCA stage had little effect on accuracy, whereas training it removed its focal occipital structure and substantially reduced accuracy. The analytical front-end is therefore best regarded as a structural prior that constrains the input of the trainable modules, rather than as an additional source of discriminative features. The t-SNE analysis further showed that the analytical stages alone produce largely class-separable features, which reduces the amount of discriminative structure that the decoder must learn from limited calibration data.

Nevertheless, several limitations remain and suggest directions for future work. First, the sub-band partition of the filter bank is hand-crafted. KDHNet learns the sub-band weights but not the sub-band boundaries, which motivates a sub-band-aware front-end whose passbands are optimized jointly with the decoder. Second, LST alignment requires a source domain large enough to estimate reliable class-wise transformations, and its benefit may decrease when the source pool is small. Stronger target-domain adaptation, for example few-shot or self-supervised alignment beyond a single linear transformation, could make inter-subject transfer more robust, particularly on low-SNR data such as BETA. Third, the design pattern studied here, a frozen closed-form prior followed by a learnable decoder, is not specific to SSVEP. Extending it to other calibration-limited BCI paradigms, such as motor imagery and event-related potentials, is a promising direction.

\section{Conclusion}

This paper presented KDHNet, a knowledge-driven hybrid network that combines a closed-form TD-RCA front-end, a learnable sub-band weighting layer, and an SSVEPformer decoder in a single end-to-end training procedure. Experiments on three public SSVEP datasets showed that KDHNet achieved the highest mean accuracy and ITR in 14 of 15 intra-subject window settings and all ten calibration-ratio settings. Under inter-subject evaluation with one calibration trial per class and LST alignment, it achieved the best performance on Datasets~I and II and the second-best performance on Dataset~III. The component-wise ablation showed that the TDCA stage was effective only when its analytical solution was kept fixed, whereas the TRCA stage and the sub-band weighting layer were robust to the choice of initialization mode. Replacing SSVEPformer with EEGNet, CCNN, or FBtCNN showed that the front-end improved every evaluated decoder on all three datasets. These results indicate that combining fixed analytical priors with trainable neural modules is an effective design for limited-calibration SSVEP decoding.
\bibliographystyle{IEEEtran} \bibliography{kdhnet}

@Article{SSVEPReview2021,
  author    = {Li, Minglun and He, Dianning and Li, Chen and Qi, Shouliang},
  journal   = {Brain Sciences},
  title     = {Brain--computer interface speller based on steady-state visual evoked potential: A review focusing on the stimulus paradigm and performance},
  year      = {2021},
  number    = {4},
  pages     = {450},
  volume    = {11},
  publisher = {MDPI},
  doi       = {10.3390/brainsci11040450},
}

@Article{DecodingReview2022,
  author  = {Yang, Man and Jung, Tzyy-Ping and Han, Jin and Xu, Minpeng and Ming, Dong},
  journal = {Journal of Biomedical Engineering},
  title   = {A review of researches on decoding algorithms of steady-state visual evoked potentials},
  year    = {2022},
  number  = {2},
  pages   = {416--425},
  volume  = {39},
  doi     = {10.7507/1001-5515.202111066},
}

@Article{Huang2023DGTransfer,
  author    = {Huang, Jiayang and Zhang, Zhi-Qiang and Xiong, Bang and Wang, Quan and Wan, Bo and Li, Fengqi and Yang, Pengfei},
  journal   = {IEEE Trans. on Neural Systems and Rehabilitation Engineering},
  title     = {Cross-subject transfer method based on domain generalization for facilitating calibration of {SSVEP}-based {BCIs}},
  year      = {2023},
  pages     = {3307--3319},
  volume    = {31},
  publisher = {IEEE},
  doi       = {10.1109/TNSRE.2023.3305202},
}

@Article{Xu2023DLSurvey,
  author    = {Xu, Dongcen and Tang, Fengzhen and Li, Yiping and Zhang, Qifeng and Feng, Xisheng},
  journal   = {Brain Sciences},
  title     = {An analysis of deep learning models in {SSVEP}-based {BCI}: A survey},
  year      = {2023},
  number    = {3},
  pages     = {483},
  volume    = {13},
  publisher = {MDPI},
  doi       = {10.3390/brainsci13030483},
}

@Article{CCA,
  author    = {Lin, Zhonglin and Zhang, Changshui and Wu, Wei and Gao, Xiaorong},
  journal   = {IEEE Trans. on Biomedical Engineering},
  title     = {Frequency recognition based on canonical correlation analysis for {SSVEP}-based {BCIs}},
  year      = {2007},
  number    = {6},
  pages     = {1172--1176},
  volume    = {54},
  publisher = {IEEE},
  doi       = {10.1109/TBME.2006.889197},
}

@Article{FBCCA,
  author    = {Chen, Xiaogang and Wang, Yijun and Gao, Shangkai and Jung, Tzyy-Ping and Gao, Xiaorong},
  journal   = {Journal of Neural Engineering},
  title     = {Filter bank canonical correlation analysis for implementing a high-speed {SSVEP}-based brain--computer interface},
  year      = {2015},
  number    = {4},
  pages     = {046008},
  volume    = {12},
  publisher = {IOP Publishing},
  doi       = {10.1088/1741-2560/12/4/046008},
}

@Article{TRCA,
  author    = {Nakanishi, Masaki and Wang, Yijun and Chen, Xiaogang and Wang, Yu-Te and Gao, Xiaorong and Jung, Tzyy-Ping},
  journal   = {IEEE Trans. on Biomedical Engineering},
  title     = {Enhancing detection of {SSVEP}s for a high-speed brain speller using task-related component analysis},
  year      = {2018},
  number    = {1},
  pages     = {104--112},
  volume    = {65},
  publisher = {IEEE},
  doi       = {10.1109/TBME.2017.2694818},
}

@Article{Liu2021TDCA,
  author    = {Liu, Bingchuan and Chen, Xiaogang and Shi, Nanlin and Wang, Yijun and Gao, Shangkai and Gao, Xiaorong},
  journal   = {IEEE Trans. on Neural Systems and Rehabilitation Engineering},
  title     = {Improving the performance of individually calibrated {SSVEP}-{BCI} by task-discriminant component analysis},
  year      = {2021},
  pages     = {1998--2007},
  volume    = {29},
  publisher = {IEEE},
  doi       = {10.1109/TNSRE.2021.3114340},
}

@Article{Chiang2021LST,
  author    = {Chiang, Kuan-Jung and Wei, Chun-Shu and Nakanishi, Masaki and Jung, Tzyy-Ping},
  journal   = {Journal of Neural Engineering},
  title     = {Boosting template-based {SSVEP} decoding by cross-domain transfer learning},
  year      = {2021},
  number    = {1},
  pages     = {016002},
  volume    = {18},
  publisher = {IOP Publishing},
  doi       = {10.1088/1741-2552/abcb6e},
}

@Article{EEGNet,
  author    = {Lawhern, Vernon J. and Solon, Amelia J. and Waytowich, Nicholas R. and Gordon, Stephen M. and Hung, Chou P. and Lance, Brent J.},
  journal   = {Journal of Neural Engineering},
  title     = {{EEGNet}: A compact convolutional neural network for {EEG}-based brain--computer interfaces},
  year      = {2018},
  number    = {5},
  pages     = {056013},
  volume    = {15},
  publisher = {IOP Publishing},
  doi       = {10.1088/1741-2552/aace8c},
}

@Article{Ding2021FBtCNN,
  author    = {Ding, Wenlong and Shan, Jianhua and Fang, Bin and Wang, Chengyin and Sun, Fuchun and Li, Xinyue},
  journal   = {IEEE Trans. on Neural Systems and Rehabilitation Engineering},
  title     = {Filter bank convolutional neural network for short time-window steady-state visual evoked potential classification},
  year      = {2021},
  pages     = {2615--2624},
  volume    = {29},
  publisher = {IEEE},
  doi       = {10.1109/TNSRE.2021.3132162},
}

@Article{CCNN,
  author    = {Ikeda, Akira and Washizawa, Yoshikazu},
  journal   = {Sensors},
  title     = {Steady-state visual evoked potential classification using complex valued convolutional neural networks},
  year      = {2021},
  number    = {16},
  pages     = {5309},
  volume    = {21},
  publisher = {MDPI},
  doi       = {10.3390/s21165309},
}

@Article{ConvCA,
  author    = {Li, Yao and Xiang, Jiayi and Kesavadas, Thenkurussi},
  journal   = {IEEE Trans. on Neural Systems and Rehabilitation Engineering},
  title     = {Convolutional correlation analysis for enhancing the performance of {SSVEP}-based brain--computer interface},
  year      = {2020},
  number    = {12},
  pages     = {2681--2690},
  volume    = {28},
  publisher = {IEEE},
  doi       = {10.1109/TNSRE.2020.3038718},
}

@Article{SSVEPNet,
  author    = {Pan, Yudong and Chen, Jianbo and Zhang, Yangsong and Zhang, Yu},
  journal   = {Journal of Neural Engineering},
  title     = {An efficient {CNN}-{LSTM} network with spectral normalization and label smoothing technologies for {SSVEP} frequency recognition},
  year      = {2022},
  number    = {5},
  pages     = {056014},
  volume    = {19},
  publisher = {IOP Publishing},
  doi       = {10.1088/1741-2552/ac8dc5},
}

@Article{SSVEPformer,
  author    = {Chen, Jianbo and Zhang, Yangsong and Pan, Yudong and Xu, Peng and Guan, Cuntai},
  journal   = {Neural Networks},
  title     = {A transformer-based deep neural network model for {SSVEP} classification},
  year      = {2023},
  pages     = {521--534},
  volume    = {164},
  publisher = {Elsevier},
  doi       = {10.1016/j.neunet.2023.04.045},
}

@Article{Deng2023TRCANet,
  author    = {Deng, Yang and Sun, Qingyu and Wang, Ce and Wang, Yijun and Zhou, S. Kevin},
  journal   = {Journal of Neural Engineering},
  title     = {{TRCA}-Net: Using {TRCA} filters to boost the {SSVEP} classification with convolutional neural network},
  year      = {2023},
  number    = {4},
  pages     = {046005},
  volume    = {20},
  publisher = {IOP Publishing},
  doi       = {10.1088/1741-2552/ace380},
}

@Article{Wei2025eTRCASbCNN,
  author    = {Wei, Qingguo and Li, Chang and Wang, Yijun and Gao, Xiaorong},
  journal   = {Scientific Reports},
  title     = {Enhancing the performance of {SSVEP}-based {BCIs} by combining task-related component analysis and deep neural network},
  year      = {2025},
  number    = {1},
  pages     = {365},
  volume    = {15},
  publisher = {Nature Publishing Group},
  doi       = {10.1038/s41598-024-84534-6},
}

@Article{Nakanishi2015,
  author    = {Nakanishi, Masaki and Wang, Yijun and Wang, Yu-Te and Jung, Tzyy-Ping},
  journal   = {PLoS One},
  title     = {A comparison study of canonical correlation analysis based methods for detecting steady-state visual evoked potentials},
  year      = {2015},
  number    = {10},
  pages     = {e0140703},
  volume    = {10},
  publisher = {Public Library of Science},
  doi       = {10.1371/journal.pone.0140703},
}

@Article{BenchmarkDataset,
  author    = {Wang, Yijun and Chen, Xiaogang and Gao, Xiaorong and Gao, Shangkai},
  journal   = {IEEE Trans. on Neural Systems and Rehabilitation Engineering},
  title     = {A benchmark dataset for {SSVEP}-based brain--computer interfaces},
  year      = {2017},
  number    = {10},
  pages     = {1746--1752},
  volume    = {25},
  publisher = {IEEE},
  doi       = {10.1109/TNSRE.2016.2627556},
}

@Article{BetaDataset,
  author    = {Liu, Bingchuan and Huang, Xiaoshan and Wang, Yijun and Chen, Xiaogang and Gao, Xiaorong},
  journal   = {Frontiers in Neuroscience},
  title     = {{BETA}: A large benchmark database toward {SSVEP}-{BCI} application},
  year      = {2020},
  pages     = {627},
  volume    = {14},
  publisher = {Frontiers Media},
  doi       = {10.3389/fnins.2020.00627},
}

@Article{Wolpaw2002,
  author    = {Wolpaw, Jonathan R. and Birbaumer, Niels and McFarland, Dennis J. and Pfurtscheller, Gert and Vaughan, Theresa M.},
  journal   = {Clinical Neurophysiology},
  title     = {Brain--computer interfaces for communication and control},
  year      = {2002},
  number    = {6},
  pages     = {767--791},
  volume    = {113},
  publisher = {Elsevier},
  doi       = {10.1016/S1388-2457(02)00057-3},
}

@Article{Rosenfeld2017,
  author  = {Rosenfeld, Jeffrey V. and Wong, Yan Tat},
  journal = {Medical Journal of Australia},
  title   = {Neurobionics and the brain--computer interface: Current applications and future horizons},
  year    = {2017},
  number  = {8},
  pages   = {363--368},
  volume  = {206},
}

@Article{wang2025mvcnet,
  author  = {Wang, Ziwei and Li, Siyang and Chen, Xiaoqing and Li, Wei and Wu, Dongrui},
  journal = {Knowledge-Based Systems},
  title   = {{MVCNet}: Multi-view contrastive network for motor imagery classification},
  year    = {2025},
  pages   = {114205},
  volume  = {328},
}

@Article{zheng2017identifying,
  author  = {Zheng, Wei-Long and Zhu, Jia-Yi and Lu, Bao-Liang},
  journal = {IEEE Trans. on Affective Computing},
  title   = {Identifying stable patterns over time for emotion recognition from {EEG}},
  year    = {2019},
  number  = {3},
  pages   = {417--429},
  volume  = {10},
}

@Article{wang2025cst,
  author  = {Wang, Ziwei and Li, Siyang and Wu, Dongrui},
  journal = {National Science Review},
  title   = {Canine {EEG} helps human: Cross-species and cross-modality epileptic seizure detection via multi-space alignment},
  year    = {2025},
  number  = {6},
  pages   = {nwaf086},
  volume  = {12},
}

@Article{bin2011high,
  author    = {Bin, Guangyu and Gao, Xiaorong and Wang, Yijun and Li, Yun and Hong, Bo and Gao, Shangkai},
  journal   = {Journal of Neural Engineering},
  title     = {A high-speed {BCI} based on code modulation {VEP}},
  year      = {2011},
  number    = {2},
  pages     = {025015},
  volume    = {8},
  publisher = {IOP Publishing},
}

@Article{bin2009online,
  author    = {Bin, Guangyu and Gao, Xiaorong and Yan, Zheng and Hong, Bo and Gao, Shangkai},
  journal   = {Journal of Neural Engineering},
  title     = {An online multi-channel {SSVEP}-based brain--computer interface using a canonical correlation analysis method},
  year      = {2009},
  number    = {4},
  pages     = {046002},
  volume    = {6},
  publisher = {IOP Publishing},
}

@InProceedings{zhang2011multiway,
  author    = {Zhang, Yu and Zhou, Guoxu and Zhao, Qibin and Onishi, Akinari and Jin, Jing and Wang, Xingyu and Cichocki, Andrzej},
  booktitle = {Int'l Conf. on Neural Information Processing},
  title     = {Multiway canonical correlation analysis for frequency components recognition in {SSVEP}-based {BCIs}},
  year      = {2011},
  pages     = {287--295},
}

@Article{pan2011enhancing,
  author    = {Pan, Jie and Gao, Xiaorong and Duan, Fang and Yan, Zheng and Gao, Shangkai},
  journal   = {Journal of Neural Engineering},
  title     = {Enhancing the classification accuracy of steady-state visual evoked potential-based brain--computer interfaces using phase constrained canonical correlation analysis},
  year      = {2011},
  number    = {3},
  pages     = {036027},
  volume    = {8},
  publisher = {IOP Publishing},
}

@Article{MSI2014,
  author    = {Zhang, Yangsong and Xu, Peng and Cheng, Kaiwen and Yao, Dezhong},
  journal   = {Journal of Neuroscience Methods},
  title     = {Multivariate synchronization index for frequency recognition of {SSVEP}-based brain--computer interface},
  year      = {2014},
  pages     = {32--40},
  volume    = {221},
  publisher = {Elsevier},
  doi       = {10.1016/j.jneumeth.2013.07.018},
}

@Article{SSCOR2019,
  author    = {Kumar, G. R. Kiran and Reddy, M. Ramasubba},
  journal   = {IEEE Trans. on Neural Systems and Rehabilitation Engineering},
  title     = {Designing a sum of squared correlations framework for enhancing {SSVEP}-based {BCIs}},
  year      = {2019},
  number    = {10},
  pages     = {2044--2050},
  volume    = {27},
  publisher = {IEEE},
  doi       = {10.1109/TNSRE.2019.2941349},
}

@Article{tanaka2013task,
  author    = {Tanaka, Hirokazu and Katura, Takusige and Sato, Hiroki},
  journal   = {NeuroImage},
  title     = {Task-related component analysis for functional neuroimaging and application to near-infrared spectroscopy data},
  year      = {2013},
  pages     = {308--327},
  volume    = {64},
  publisher = {Elsevier},
}

@Article{sun2021similarity,
  author    = {Sun, Qiang and Chen, Minyou and Zhang, Li and Li, Changsheng and Kang, Wenfa},
  journal   = {Journal of Neural Engineering},
  title     = {Similarity-constrained task-related component analysis for enhancing {SSVEP} detection},
  year      = {2021},
  number    = {4},
  pages     = {046080},
  volume    = {18},
  publisher = {IOP Publishing},
}

@Article{Waytowich2018,
  author    = {Waytowich, Nicholas and Lawhern, Vernon J. and Garcia, Javier O. and Cummings, Jennifer and Faller, Josef and Sajda, Paul and Vettel, Jean M.},
  journal   = {Journal of Neural Engineering},
  title     = {Compact convolutional neural networks for classification of asynchronous steady-state visual evoked potentials},
  year      = {2018},
  number    = {6},
  pages     = {066031},
  volume    = {15},
  publisher = {IOP Publishing},
  doi       = {10.1088/1741-2552/aae5d8},
}

@Article{Guney2022,
  author    = {G{\"u}ney, Osman Berke and Oblokulov, Muhtasham and Ozkan, Huseyin},
  journal   = {IEEE Trans. on Biomedical Engineering},
  title     = {A deep neural network for {SSVEP}-based brain--computer interfaces},
  year      = {2022},
  number    = {2},
  pages     = {932--944},
  volume    = {69},
  publisher = {IEEE},
  doi       = {10.1109/TBME.2021.3110440},
}

@Article{song2022eeg,
  author    = {Song, Yonghao and Zheng, Qingqing and Liu, Bingchuan and Gao, Xiaorong},
  journal   = {IEEE Trans. on Neural Systems and Rehabilitation Engineering},
  title     = {{EEG} {C}onformer: Convolutional {T}ransformer for {EEG} decoding and visualization},
  year      = {2023},
  pages     = {710--719},
  volume    = {31},
  publisher = {IEEE},
}

@Article{wu2020transfer,
  author  = {Wu, Dongrui and Xu, Yifan and Lu, Bao-Liang},
  journal = {IEEE Trans. on Cognitive and Developmental Systems},
  title   = {Transfer learning for {EEG}-based brain--computer interfaces: A review of progress made since 2016},
  year    = {2022},
  number  = {1},
  pages   = {4--19},
  volume  = {14},
}

@Article{wang2025dbconformer,
  author  = {Ziwei Wang and Hongbin Wang and Tianwang Jia and Xingyi He and Siyang Li and Dongrui Wu},
  journal = {IEEE Journal of Biomedical and Health Informatics},
  title   = {{DBConformer}: Dual-branch convolutional {Transformer} for {EEG} decoding},
  year    = {2026},
  number  = {5},
  pages   = {4134--4147},
  volume  = {30},
}

@Article{Wang2023TASA,
  author  = {Wang, Ziwei and Zhang, Wen and Li, Siyang and Chen, Xinru and Wu, Dongrui},
  journal = {Journal of Neural Engineering},
  title   = {Unsupervised domain adaptation for cross-patient seizure classification},
  year    = {2023},
  number  = {6},
  pages   = {066002},
  volume  = {20},
}

@Article{wang2026ckd,
  author    = {Wang, Ziwei and He, Xingyi and Wang, Hongbin and Wu, Dongrui},
  journal   = {IEEE Trans. on Biomedical Engineering},
  title     = {{CKD}: Contrastive knowledge distillation for cross-dataset {EEG} classification},
  year      = {2026},
  note      = {Early access},
  publisher = {IEEE},
}

@Article{Wilcoxon1945,
  author  = {Wilcoxon, Frank},
  journal = {Biometrics Bulletin},
  title   = {Individual comparisons by ranking methods},
  year    = {1945},
  number  = {6},
  pages   = {80--83},
  volume  = {1},
  doi     = {10.2307/3001968},
}

@Article{Holm1979,
  author  = {Holm, Sture},
  journal = {Scandinavian Journal of Statistics},
  title   = {A simple sequentially rejective multiple test procedure},
  year    = {1979},
  number  = {2},
  pages   = {65--70},
  volume  = {6},
}

\end{document}